\pdfoutput=1
\documentclass[final]{arxiv}
\usepackage{times}
\usepackage{epsfig}
\usepackage{graphicx}
\usepackage{amsmath}
\usepackage{amssymb}

\usepackage[switch]{lineno}
\usepackage[ruled,vlined]{algorithm2e}
\usepackage{bm}
\usepackage{dsfont}

\newcommand{\mb}{\mathbf}
\usepackage{verbatim}

\usepackage{amsmath}
\usepackage{subfigure}
\usepackage{mathtools}  
\usepackage{amsfonts}
\usepackage{tabularx}

\usepackage[pagebackref=true,breaklinks=true,colorlinks,bookmarks=false]{hyperref}
\begin{document}

\title{DSRNA: Differentiable Search of Robust Neural Architectures}

\author{Ramtin Hosseini\\
UC San Diego\\
9500 Gilman Dr, La Jolla, CA 92093\\
{\tt\small rhossein@ucsd.edu}

\and
Xingyi Yang\\
UC San Diego\\
9500 Gilman Dr, La Jolla, CA 92093\\
{\tt\small x3yang@eng.ucsd.edu}

\and

Pengtao Xie\\
UC San Diego\\
9500 Gilman Dr, La Jolla, CA 92093\\
{\tt\small p1xie@eng.ucsd.edu}

}

\maketitle

\begin{abstract}

   In deep learning applications, the architectures of deep neural networks are crucial in achieving high accuracy. Many methods have been proposed to search for high-performance neural architectures automatically. However, these searched architectures  are prone to adversarial attacks. A small perturbation of the input data can render the architecture to change prediction outcomes significantly. To address this problem, we propose methods to perform differentiable search of robust neural architectures. In our methods, two differentiable metrics are defined to measure architectures' robustness , based on certified lower bound and Jacobian norm bound. Then we search for robust architectures by maximizing the robustness metrics. Different from previous approaches which aim to improve architectures' robustness in an implicit way: performing adversarial training and injecting random noise, our methods explicitly and directly maximize robustness metrics to harvest robust architectures. 
    On CIFAR-10, ImageNet, and  MNIST, we perform game-based evaluation and verification-based evaluation on the robustness of our methods. The experimental results show that our methods 1) are more robust to various norm-bound attacks than several robust NAS baselines; 2) are more accurate than baselines when there are no attacks; 3) have significantly higher certified lower bounds than baselines.  
\end{abstract}

\section{Introduction}

In deep learning applications, the architectures of neural models play a crucial role in improving performance. For example, on the ImageNet~\cite{imagenet_cvpr09} benchmark, the image classification error is reduced from 16.4\% to 3.57\%, when the architecture is evolved from AlexNet~\cite{krizhevsky2012imagenet} to ResNet~\cite{he2016deep}. Previously, neural architectures are mostly designed by humans, which is time-consuming to obtain a highly-performant architecture. Recently, automated neural architecture search~\cite{zoph2016neural,zoph2018learning,real2019regularized,real2017large,weng2019unet,weng2019automatic} which develops  algorithms to find out the optimal architecture that yields the best performance on the validation datasets, has raised much attention and achieved promising results. For example, on the CIFAR-10 dataset, an automatically searched architecture~\cite{liu2018darts} achieves an image classification error rate of 2.76\% while the error achieved by state-of-the-art human-designed architecture is 3.46\%.

As we will show in the experiments, the architectures searched by existing methods are prone to adversarial attacks. A small perturbation (which is not perceivable by humans) of the input data can render the architecture to change prediction outcomes significantly. Many approaches~\cite{goodfellow2014explaining,carlini2017towards,madry2017towards,cohen2019certified,lecuyer2019certified} have been proposed to improve the robustness of DNNs. In these approaches, the architecture of a DNN is provided by humans, and the defense method focuses on training the weights in this architecture in a robust way. However, the robustness of a DNN is not only relevant to its weight parameters, but also determined by the architecture. It is important to search for architectures that are robust to adversarial attacks as well. 

In this paper, we develop a novel approach for robust NAS. We define two differentiable metrics to measure the robustness of architectures and formulate robust NAS as an optimization problem that aims to find out an optimal architecture by maximizing  the robustness metrics. The first metric is defined based on certified lower bound~\cite{boopathy2019cnn}. Linear bounding methods are applied to individual building blocks in the differentiable architecture search space and these individual bounds are composed to obtain global bounds for the entire neural architecture. The second metric is based on the Jacobian norm bound, where the robustness is measured by how much the output shifts as the input is perturbed. The shift is upper bounded by the norms of row vectors in the Jacobian matrix of the neural architecture. Our approach is applicable to various forms of differentiable architecture search methods (e.g., DARTS~\cite{liu2018darts},  PC-DARTS \cite{xu2019pc}, P-DARTS \cite{chen2019progressive}, and DARTS+ \cite{liang2019darts+}) and is robust against adversarial attacks in various norm choices. Previously, robust NAS has been investigated in \cite{guo2020meets,chen2020stabilizing}, based on adversarial training of randomly sampled sub-architectures~\cite{guo2020meets} and differentiable architecture variables~\cite{chen2020stabilizing}. Unlike these methods that achieve robustness implicitly via adversarial training, our method explicitly defines robustness metrics and directly optimizes these metrics to obtain robust architectures.

On CIFAR-10, ImageNet, and MNIST, we perform game-based evaluation and verification-based evaluation on the robustness of our methods. The experimental results show that our methods 1) are more robust to various norm-bound attacks  than several robust NAS baselines; 2) are more accurate than baselines when there are no attacks; 3) have significantly higher certified lower bounds than baselines.

The major contributions of this paper include:
\begin{itemize}
\item We propose a novel robust NAS method, which searches robust architectures by maximizing differentiable robustness metrics, defined based on certified lower bound and Jacobian norm bound. Our methods have strong guarantees in obtaining robust architectures by explicitly and directly maximizing robustness measures. In contrast, previous approaches perform implicit robustification of architectures via adversarial training, which is not guaranteed to yield robust architectures. Besides, our methods can be applied to robustify any differentiable NAS methods, in a principled and unified way.  

\item Experiments on ImageNet, CIFAR-10, and MNIST show that the architectures searched by our methods are robust to various forms of adversarial attacks and are as accurate as state-of-the-art NAS methods when there are no attacks. Our methods are consistently more robust than previous approaches against various attacks. In contrast,  previous approaches are effective for certain type of attacks, but ineffective for other types.   

\end{itemize}
The rest of the paper is organized as follows. Section \ref{sec:2} reviews related works. Section \ref{sec:3} and \ref{sec:4} present the method and experiments. Section \ref{sec:5} concludes the paper.

\section{Related Works}
\label{sec:2}

\subsection{Neural Architecture Search}
 In general, there are three paradigms of methods in NAS: reinforcement learning (RL) approaches~\cite{zoph2016neural,pham2018efficient,zoph2018learning}, evolutionary learning approaches~\cite{liu2017hierarchical,real2019regularized}, and differentiable approaches~\cite{cai2018proxylessnas,liu2018darts,xie2018snas}. In RL-based approaches, a policy is learned to iteratively generate new architectures by maximizing a reward, which is the accuracy on the validation set. Evolutionary learning approaches represent the architectures as individuals in a population. Individuals with high fitness scores (validation accuracy) have the privilege to generate offsprings, which replace individuals with low fitness scores. Differentiable NAS approaches adopt a network pruning strategy. On top of an over-parameterized network, the weights of connections between nodes are learned using gradient descent. Then weights close to zero are later pruned. There have been many efforts devoted to improving differentiable NAS methods. 
 In P-DARTS \cite{chen2019progressive}, the depth of searched architectures is allowed to grow progressively during the training process. 
 
  Search space approximation and
regularization approaches are developed to reduce computational overheads and improve search stability. 

  PC-DARTS \cite{xu2019pc} reduces the redundancy in exploring the search space by sampling a small portion  of a super network. Operation search is performed in a subset of channels with the held out part bypassed in a shortcut.

 DARTS+ \cite{liang2019darts+} leverages early stopping to avoid the collapse of DARTS' performance.

\subsection{Adversarial Attacks and Defenses}
Adversarial attacks aim to perturb  input data examples by adding imperceptible noises so that the prediction results are altered significantly.  
In white-box attack, the adversary has full access to the target model \cite{szegedy2013intriguing,chen2017show,cheng2020seq2sick,zugner2018adversarial}, while in the black-box attack, the target model is unknown to the adversary \cite{chen2017zoo,ilyas2018black,tu2019autozoom,cheng2018query}. In targeted attacks, the adversary aims to change the prediction outcome in certain classes, while untargeted attacks are not class-specific.  
Arguably, the most popular and effective white-box untargeted attacks with various norm-bounds are: fast gradient sign method (FGSM) \cite{goodfellow2014explaining}, projected gradient descent (PGD) \cite{madry2017towards}, and Carlini \& Wagner (C\&W) \cite{carlini2017towards}. FGSM is a  single step attack algorithm that aims to increase the adversarial loss by updating its gradient sign. 
PGD is a more general version of FGSM that runs over several iterations to increase the adversarial loss. 
The attacks of FGSM and PGD are based on $l_{\infty}$-norm bound, while those in C\&W are based on  $l_{0}$, $l_{2}$, and $l_{\infty}$ norms. C\&W is particularly effective for $l_{2}$-norm attacks.

To improve the robustness of  neural networks against adversarial attacks, many adversarial defense methods have been proposed, 
such as random smoothing \cite{lecuyer2019certified,cohen2019certified} and adversarial training \cite{goodfellow2014explaining,madry2017towards,carlini2017towards}.
Most of these defense methods assume the neural architectures are manually designed by human and focus on improving the robustness of network weights. Automatically searching for robust architectures is largely under-explored. In \cite{devaguptapu2020empirical},  experiments show that architectures searched by existing NAS methods such as DARTS, PC-DARTS, and P-DARTS are vulnerable to various forms of  adversarial attacks. 
 To address this issue, studies have been conducted to robustify NAS methods. RobNet \cite{guo2020meets}
 used one-shot NAS to obtain a large number of networks and then 
 studied the patterns of  architectures that are robust against adversarial attacks. They discovered that using dense connectivity and  adding convolution operations to direct connection edges help to improve robustness. Chen et al.~\cite{chen2020stabilizing} proposed performing adversarial training and random smoothing on architecture variables, which can improve the robustness of DARTS-based methods. Our work takes a different approach for robustifying architectures, where we explicitly define differentiable metrics to measure architectures' robustness and search for robust architectures by maximizing these metrics.



\subsection{Robustness Verification of Neural Networks}
\label{sec:2.3}
Robustness verification aims to provide certified defense against any possible attacks under a
threat model. A robustness certificate $\epsilon$ means the prediction outcome cannot be changed if the strength of the attack is smaller than $\epsilon$. Many verification approaches \cite{wong2018provable,weng2018towards,zhang2018efficient,dvijotham2018dual} have been proposed, which focus on achieving tighter lower bounds of the robustness certificate, computing bounds for various complex building blocks in neural networks, and improving the efficiency in computing the bounds. Dvijotham et al.~\cite{dvijotham2018dual} formulate verification as an optimization problem and seek bounds of the certificate by solving a Lagrangian relaxation of the optimization problem. Weng et al.~\cite{weng2018towards} propose methods to verify the  robustness of Rectified Linear Unit (ReLU) networks by bounding the ReLU units with  linear functions or local Lipschitz constant. CNN-Cert~\cite{boopathy2019cnn} applies linear bounding techniques to provide certified lower bounds for various operations including  convolution, pooling, batch
normalization, residual blocks,   activation
functions, etc.

\section{Methods}
\label{sec:3}

We begin with defining  differentiable metrics to measure the robustness of neural architectures. Then we propose a robust NAS framework that performs optimization in the architecture search space to maximize the robustness metrics. The objective function explores a tradeoff between predictive accuracy and robustness and  can be efficiently optimized using gradient-based methods.  

\subsection{Defining Differentiable  Robustness Metrics}

In this section, we define two differentiable metrics to measure the robustness of neural architectures. 
The first one is based on robustness certification methods~\cite{boopathy2019cnn}. Specifically, given an architecture, we seek to obtain a certified lower bound of this architecture and use the bound to measure robustness. The architecture with the larger lower bound is more robust against different attacks. The second metric is based on upper-bounding the shift of the model's prediction when the inputs are perturbed, and the bound is based on the norm of the Jacobian matrix of the architecture. The smaller the upper bound is, the more robust the network is.

\subsubsection{Measuring Robustness Based on Certified Bound}
\label{sec:cb}

One way to measure the robustness of a neural network is to use the verified robustness certificate. A certificate with value $\epsilon(\mb{x})$ means that the model prediction on the input data $\mb{x}$ cannot be changed if the attack strength is smaller
than $\epsilon(\mb{x})$. A larger $\epsilon(\mb{x})$ indicates more robustness. In practice, it is infeasible to obtain the exact robustness certificate of a model. Instead, one can derive lower bounds of $\epsilon(\mb{x})$ and use these lower bounds as surrogates for measuring robustness. Given an architecture search space comprised of various building blocks such as ReLU-Conv-BN, (dilated) separable convolutions, pooling operations, etc., we perform linear bounding~\cite{boopathy2019cnn} on these building blocks and compose the individual bounds to obtain a certified lower bound for each architecture in the search space. These bounds are differentiable functions of architecture variables and are amenable for gradient-based optimization. In the sequel, we discuss how to derive the certified upper and lower bounds for each type of building blocks.

\paragraph{ReLU-Conv-BN Block}
 The ReLU-Conv-BN building block consists of three consecutive operations including rectified linear unit (ReLU) as a nonlinear activation operation, convolution, and batch normalization (BN). Let $\Phi^{r}$ and $\Phi^{r-2}$ be the output and input of an ReLU-Conv-BN block $r$, then we have  
 \begin{equation}
 \label{eq:11}
\Phi^{r-1} = W^{r-1} * \sigma(\Phi^{r-2}) + b^{r-1}
\end{equation}
\begin{equation}
\label{eq:12}
\Phi^{r} = \gamma_{bn}\frac{\Phi^{r-1}-\mu_{bn}}{\sqrt{\sigma^{2}_{bn}+\epsilon_{bn}}}+\beta_{bn}
\end{equation}
where $\sigma(\cdot)$ is the ReLU function. $W^{r-1}$ and $b^{r-1}$ are the weight parameters and bias parameters in the convolution operation. $\mu_{bn}$ and $\sigma^{2}_{bn}$ are the mean and variance of a batch of $\Phi^{r-1}$ in batch normalization. $\gamma_{bn}$, $\epsilon_{bn}$, and $\beta_{bn}$ are hyperparameters in BN.

By applying linear bounds to these equations, we get these upper and lower bounds:
\begin{equation}
\label{eq:13}
A_{L,bn}^{r} * \Phi^{r-1} + B_{L,bn}^{r} \leq \Phi^{r} \leq A_{U,bn}^{r} * \Phi^{r-1} + B_{U,bn}^{r}
\end{equation}
\begin{equation}
\label{eq:14}
 A_{L,bn}^{r}  \Phi^{r-1} + B_{L,bn}^{r} \geq  A_{L,bn}^{r}  (A_{L,conv}^{r-1} \Phi^{r-2}+B_{L,conv}^{r-1}) + B_{L,bn}^{r}
\end{equation}
\begin{equation}
\label{eq:15}
 A_{U,bn}^{r}  \Phi^{r-1} + B_{U,bn}^{r} \leq  A_{U,bn}^{r}  (A_{U,conv}^{r-1} \Phi^{r-2}+B_{U,conv}^{r-1}) + B_{U,bn}^{r}
\end{equation}
where $A_{L,bn}$, $A_{U,bn}$, $B_{L,bn}$, and $B_{U,bn}$ are constants that can be computed as in  \cite{boopathy2019cnn}:
\begin{equation}
\label{eq:16}
A_{L,bn}^{r} = A_{U,bn}^{r} = \frac{\gamma _{bn}}{\sqrt{\sigma^{2}_{bn}+\epsilon_{bn}}}
\end{equation}
\begin{equation}
\label{eq:17}
B_{L,bn}^{r} = B_{U,bn}^{r} = \frac{-\gamma _{bn} \mu_{bn}}{\sqrt{\sigma^{2}_{bn}+\epsilon_{bn}}}+\beta_{bn}
\end{equation}
and $A_{L,conv}$, $A_{U,conv}$, $B_{L,conv}$, $B_{U,conv}$ are constant tensors.

\paragraph{(Dilated) Separable Convolutions} Another two types of building blocks in our search space are separable convolutions and dilated separable convolutions. Dilated separable convolutions consist of four consecutive operations: ReLU, convolution, convolution, and batch normalization (BN). Separable convolutions consist of two consecutive dilated separable convolutions.  

Let $\Phi^{r-3}$ and $\Phi^{r}$ denote the input and output of a dilated separable convolution, then:
\begin{equation}
\label{eq:18}
\Phi^{r-1} = W^{r-1} * (W^{r-2} * \sigma(\Phi^{r-3}) + b^{r-2}) + b^{r-1}
\end{equation}
where $W^{r-1}$ and $W^{r-2}$ are weights of convolutions; $b^{r-2}$ and $b^{r-1}$ are bias parameters in convolutions. The calculation of $\Phi^{r}$ is the same as that in Eq.(\ref{eq:12}). 

We can again use Eq.(\ref{eq:13}) to find the upper and lower bound of $\Phi^{r}$, which are:

 \begin{equation}
 \label{eq:20}
 \begin{array}{l}
 A_{L,bn}^{r} * \Phi^{r-1} + B_{L,bn}^{r} \geq  
 A_{L,bn}^{r} * (A_{L,conv}^{r-1} *(W^{r-2} \\* \Phi^{r-3}+b^{r-2})+B_{L,conv}^{r-1}) + B_{L,bn}^{r}
  \end{array}
\end{equation}

\begin{equation} 
\label{eq:21}
 \begin{array}{l}
 A_{U,bn}^{r} * \Phi^{r-1} + B_{U,bn}^{r} \leq 
 A_{U,bn}^{r} * (A_{U,conv}^{r-1} *(W^{r-2} \\* \Phi^{r-3}+b^{r-2})+B_{U,conv}^{r-1}) + B_{U,bn}^{r}
  \end{array}
\end{equation}
The upper and lower bound for separable convolution operations can be derived in a similar way.

\paragraph{Pooling Operations}
Let $\Phi^{r-1}$ and $\Phi^{r}$ denote the input and output of a pooling operation $r$. We have the following lower and upper bound of $\Phi^{r}$:
 \begin{equation}
 \label{eq:23}
 \begin{split}
A_{L,pool}^{r} * \Phi^{r-1} + B_{L,pool}^{r} \leq \Phi^{r}
\leq A_{U,pool}^{r} * \Phi^{r-1} + B_{U,pool}^{r}
 \end{split}
\end{equation}

\paragraph{Robustness Metric} Given the lower and upper bounds of individual building blocks, we are ready to derive a certified lower bound for the entire network as a measure of the robustness of its architecture. In differentiable architecture search~\cite{liu2018darts}, the neural network is overparameterized with many building blocks that are organized into a directed acyclic graph (DAG). The output of each block is multiplied with a positive scalar. The larger the scalar is, the more critical the block is. After learning, a subset of blocks with the largest scalars are selected to form the final architecture of this network. Therefore, these scalars (called architecture variables) represent the architecture. Given a block with lower bound $L$ and upper bound $U$, after multiplying with an architecture variable $\alpha$, this block has a lower bound of $\alpha L$ and $\alpha U$. Following the topological order of blocks in the DAG, we recursively compose the lower and upper bounds (multiplied with architecture variables) of blocks and get a global lower and upper bound for the entire network. These two bounds are functions of architecture variables and the input data example. The lower bound is used as the robustness metric.

\subsubsection{Measuring Robustness with Jacobian Regularization}

When the architecture search space is large, computing gradients of the certified lower bound with respect to architecture variables is time-consuming. To address this problem, we investigate another measure of robustness, which is computationally efficient. Let $f(\mb{x})$ denote the neural network which takes a data example $\mb{x}\in\mathbb{R}^D$ as input and outputs a $K$-dimensional vector. Similar to the robustness metric defined in Section~\ref{sec:cb}, the architecture search space is differentiable, where continuous architecture variables are multiplied to the outputs of building blocks. Therefore, $f(\mb{x})$ is a continuous function of the architecture variables. Let $\mb{x}+\boldsymbol\epsilon$ be an adversarial example where $\boldsymbol\epsilon$ is a small perturbation vector. We assume the $p$-norm  of $\boldsymbol\epsilon$ is less equal to a small scalar $\delta$: $\|\boldsymbol\epsilon\|_p\leq \delta$. We measure the robustness of the network using the following quantity:
\begin{equation}
  S=  -\mathbb{E}_{\mb{x}}\mathbb{E}_{\boldsymbol\epsilon}\left[\frac{1}{K} \sum_{k=1}^K|f_k(\mb{x}+\boldsymbol\epsilon)-f_k(\mb{x})|\right]
\end{equation}
where $a=1/K \sum_{k=1}^K|f_k(\mb{x}+\boldsymbol\epsilon)-f_k(\mb{x})|$ is the average change of the output across all dimensions when $\mb{x}$ is perturbed with $\boldsymbol\epsilon$ and $S$ is the expectation of $a$ defined with respect to the distributions of $\mb{x}$ and $\boldsymbol\epsilon$. The smaller this quantity is, the more robust the network is: intuitively, a network is robust if for every input data example, no matter how it is perturbed, the change of network output is small. According to Taylor expansion, we have:

\begin{equation}
  f_k(\mb{x}+\boldsymbol\epsilon)- f_k(\mb{x}) \approx \left[\frac{\partial f_k(\mb{x})}{\partial \mb{x}}\right]^\top \boldsymbol\epsilon
\end{equation}
Let $\mb{J}(\mb{x})$ denote the Jacobian matrix at $\mb{x}$ where $J_{kj}=\partial f_k(\mb{x})/\partial x_j$. Then $\partial f_k(\mb{x})/\partial \mb{x}=\mb{J}_k(\mb{x})$ where $\mb{J}_k(\mb{x})$ is the $k$-th row vector of  $\mb{J}(\mb{x})$. According to Hölder's inequality, we have: 
\begin{equation}
    \left|\mb{J}_k(\mb{x})^\top \boldsymbol\epsilon\right|\leq \left\Vert\mb{J}_k(\mb{x})\right\Vert_q\|\epsilon\|_p\leq \left\Vert\mb{J}_k(\mb{x})\right\Vert_q \delta
\end{equation}
where $\frac{1}{p}+\frac{1}{q}=1$.

Putting these pieces together, we have:
\begin{equation}
\begin{array}{l}
    -\mathbb{E}_{\mb{x}}\mathbb{E}_{\boldsymbol\epsilon}\left[\frac{1}{K} \sum_{k=1}^K|f_k(\mb{x}+\boldsymbol\epsilon)-f_k(\mb{x})|\right]\\
    \approx -\mathbb{E}_{\mb{x}}\mathbb{E}_{\boldsymbol\epsilon}\left[\frac{1}{K} \sum_{k=1}^K |\mb{J}_k(\mb{x})^\top \boldsymbol\epsilon|\right]\\
    \geq -\mathbb{E}_{\mb{x}}\mathbb{E}_{\boldsymbol\epsilon}\left[\frac{1}{K} \sum_{k=1}^K \left\Vert\mb{J}_k(\mb{x})\right\Vert_q \delta \right]\\
    =-\delta\mathbb{E}_{\mb{x}}\left[\frac{1}{K} \sum_{k=1}^K \left\Vert\mb{J}_k(\mb{x})\right\Vert_q  \right]\\
    \approx -\frac{\delta}{N}\sum_{i=1}^N\left[\frac{1}{K} \sum_{k=1}^K \left\Vert\mb{J}_k(\mb{x}_i)\right\Vert_q  \right]
    \end{array}
\end{equation}
where in the last step, the expectation is approximated by the mean on a set of data examples $\{\mb{x}_i\}_{i=1}^N$. To maximize $S$ for achieving robustness, we can maximize its approximated lower bound $-\delta/N\sum_{i=1}^N\left[1/K \sum_{k=1}^K \left\Vert\mb{J}_k(\mb{x}_i)\right\Vert_q  \right]$. We call it the Jacobian norm bound. It is a function of the  architecture variables. 
For $l_2$ and $l_\infty$ norm bound attacks, $\sum_{k=1}^K \left\Vert\mb{J}_k(\mb{x})\right\Vert_q$ is the  Frobenius norm and $l_1$ norm of the Jacobian matrix respectively. We use the method in \cite{hoffman2019robust} to compute the Jacobian matrix efficiently based on random projection.  

\begin{table*}[t]
\begin{center}
  \begin{tabular}{|l||c|c|c|c|c|}
\hline
Method &  PGD (10) & PGD (20) & PGD (100) & FGSM & C\&W\\
\hline
\hline

RobNet-large \cite{guo2020meets}  & 49.49 & 49.44 & 49.24 & 54.98 &47.19\\

RobNet-free \cite{guo2020meets} & 52.80 & 52.74 & 52.57 & 58.38 &46.95\\
\hline
SDARTS-ADV \cite{chen2020stabilizing}$^{\ast}$  &  56.94 $\pm$ 0.02 & 56.90 $\pm$ 0.04 & 56.77 $\pm$ 0.17 & 63.84 $\pm$ 0.02 & 42.67 $\pm$ 0.09\\
PC-DARTS-ADV \cite{chen2020stabilizing}$^{\ast}$  &  57.15 $\pm$ 0.02& 57.11 $\pm$ 0.05& 56.83 $\pm$  0.21& 65.29 $\pm$ 0.03& 42.58 $\pm$ 0.04\\
\hline
DSRNA-CB (ours)$^{\ast}$  &\textbf{60.31 $\pm$ 0.07} & \textbf{60.22 $\pm$ 0.11} & \textbf{59.93 $\pm$ 0.24} & \textbf{69.88 $\pm$ 0.09} &\textbf{63.01 $\pm$ 0.07} \\

DSRNA-Jacobian (Ours)$^{\ast}$  &  59.81 $\pm$ 0.02 & 59.77 $\pm$ 0.04 & 59.47 $\pm$ 0.14& 68.92 $\pm$ 0.02 &62.87 $\pm$  0.04\\

\hline
\end{tabular}
   \end{center}

\caption{Accuracy (\%) (standard deviation) of different  methods under 
   various norm-bound attacks on CIFAR-10. $^{\ast}$Average of five  runs. 
   }\label{tab:title2}
\end{table*}

\begin{table*}[hbt!]
\begin{center}

  \begin{tabular}{|l||c|c|c|c|}
\hline
\textbf{Method} & \textbf{Test Acc. (\%)} & \vtop{\hbox{\strut \textbf{Params}}\hbox{\strut \textbf{(M)}}}& \vtop{\hbox{\strut \textbf{Search Cost}}\hbox{\strut \textbf{(GPU days)}}}& \vtop{\hbox{\strut \textbf{Search}}\hbox{\strut \textbf{Method}}}\\
\hline
\hline
NASNet-A \cite{zoph2016neural}& 97.35 & 3.3 &1800 &RL\\

AmoebaNet-B \cite{real2019regularized}& 97.45 &2.8  &3150 &evolution\\

PNAS \cite{liu2018progressive}$^{\dagger}$ & 96.59 &3.2 &255 &SMBO\\

ENAS \cite{pham2018efficient}& 97.11 &4.6 &0.5 &RL\\

\hline
DARTS (1st) \cite{liu2018darts} & 97.00 $\pm$ 0.14 &3.3 &1.5 &gradient\\

DARTS (2nd) \cite{liu2018darts}& 97.26 $\pm$ 0.09 & 3.3&4.0 &gradient\\

SNAS (moderate) \cite{xie2018snas}& 97.15 &2.8 &1.5 &gradient\\

ProxylessNAS \cite{cai2018proxylessnas}$^{\ast}$ & 97.92  &--&4.0 &gradient\\

ASAP \cite{noy2020asap}& \textbf{98.01} &2.5 &0.2 &gradient\\

R-DARTS (L2) \cite{zela2019understanding} & 97.05 $\pm$ 0.21&-- &1.6 &gradient\\

DARTS+ \cite{liang2019darts+} & 97.68 &3.7&0.4 &gradient\\

P-DARTS \cite{chen2019progressive} & 97.50 &3.4 &0.3 &gradient\\

PC-DARTS \cite{xu2019pc} & 97.43 $\pm$ 0.07 & 3.6 &0.1 &gradient\\

\hline
RobNet-large \cite{guo2020meets} & 78.57 &6.9 &-- &one shot\\
RobNet-free \cite{guo2020meets} & 82.79 &5.5 &-- & one shot\\
\hline
SDARTS-RS  \cite{chen2020stabilizing} & 97.33 $\pm$ 0.03 &3.4&0.4 &gradient\\
SDARTS-ADV \cite{chen2020stabilizing} & 97.39 $\pm$ 0.02  &3.3&1.3 &gradient\\
PC-DARTS-ADV \cite{chen2020stabilizing} & 97.51 $\pm$ 0.04 &3.5&0.4 &gradient\\
\hline
DSRNA-CB (ours)$^{\ddagger}$ & 97.42 $\pm$ 0.07 &3.5 & 4.0 &gradient \\
DSRNA-Jacobian (ours)$^{\ddagger}$ & 97.50 $\pm$ 0.03 &3.5 & 1.0 &gradient\\
\hline
\end{tabular}
\end{center}
  \caption{Accuracy (\%) (mean and standard deviation) of different NAS methods when there are no attacks. $^{\ddagger}$Average of five runs. $^{\dagger}$Training without cutout augmentation. $^{\ast}$Using a different search space.
  } \label{tab:title1} 
  \vspace{-0.3cm}
\end{table*}

\subsection{Differentiable Search of Robust Neural Architectures}
Given the robustness metrics defined based on certified lower bound and Jacobian norm bound,  which are  increasing functions of the architecture variables (i.e.,   larger values of the metrics indicate that the architecture is more robust), we search for robust architectures by maximizing these robustness metrics. The formulation is as follows: 
\begin{equation}
\label{eq:5}
\begin{array}{ll}
  \min_{\alpha}   & \sum\limits_{i=1}^{M} L(w^{*}(\alpha),\alpha, x^{(\textrm{val})}_i) -\gamma R(w^{*}(\alpha),\alpha, x^{(\textrm{val})}_i) \\
   s.t.   & w^{*}(\alpha) = \textrm{argmin}_{w} \;\; \sum\limits_{i=1}^{N} L(w,\alpha, x^{(\textrm{tr})}_i)
\end{array}
\end{equation}
where $\alpha$ denotes the set of architecture variables, and $w$ denotes the weight parameters of blocks. $R$ denotes the robustness metric (either based on certified lower bound or Jacobian norm bound). $M$ is the number of validation examples, and $N$ is the number of training examples. On each validation example $x^{(\textrm{val})}_i$, we measure the robustness $R$ and predictive loss $L$ of the architecture $\alpha$ and aim to search for an optimal architecture that yields the largest robustness and smallest predictive loss on the validation set. $\gamma$ is a tradeoff parameter balancing these two objectives.  Similar to \cite{liu2018darts}, this is a bi-level optimization problem. 
In the inner optimization problem, given an architecture configuration $\alpha$, an optimal set of weights $w^{*}(\alpha)$ is learned by minimizing the training loss $\sum_{i=1}^{N} L(w,\alpha, x^{(\textrm{tr})}_i)$. Note that $w^{*}(\alpha)$ is a function of $\alpha$: each architecture configuration $\alpha$ corresponds to a set of optimal weights $w^{*}(\alpha)$. $w^{*}(\alpha)$ and $\alpha$ are both used to measure the robustness and predictive loss on the validation set. 
In the outer optimization problem, we learn the architecture variables by minimizing the validation loss and maximizing the robustness metric, i.e., searching for an  architecture that is accurate and robust. When $R$ is the metric based on certified bound (CB), our method is denoted as DSRNA-CB; when $R$ is the metric based on Jacobian norm bound, our method is denoted as DSRNA-Jacobian. 
The algorithm for solving the optimization problem in Eq.(16) can be derived in a similar way to that in DARTS~\cite{liu2018darts}. We approximate $w^*(\alpha)$ using one step gradient descent update of $w$ with respect to the training loss. Then we plug in this approximation into the validation loss and robustness metric, and perform gradient descent update of $\alpha$ with respect to the approximated objective in the first line in Eq.(16). The detailed algorithm is deferred to  the supplements.

\begin{table*}[t]
\begin{center}

\begin{tabular}{|l||c|c|c|c|}
\hline
Method & Without attack & PGD (100)  & FGSM & C\&W\\
  \hline
  \hline
RobNet-large~\cite{guo2020meets} & 61.26 & 37.14 & 39.74 & 25.73\\
\hline
SDARTS-ADV~\cite{chen2020stabilizing} $^{\ast}$ & 74.85  $\pm$ 0.06& 46.54 $\pm$ 0.13& 48.09 $\pm$ 0.07& 36.86 $\pm$ 0.10\\
PC-DARTS-ADV~\cite{chen2020stabilizing}  $^{\ast}$ & 75.73 $\pm$ 0.07 & \textbf{46.59 $\pm$ 0.15}& 48.25 $\pm$ 0.08 & 36.69 $\pm$ 0.09\\
\hline
DSRNA-CB (ours)$^{\ast}$  & 75.84 $\pm$ 0.11 & 45.39 $\pm$ 0.18& \textbf{50.89 $\pm$ 0.07} & \textbf{43.64 $\pm$ 0.19}\\
DSRNA-Jacobian (ours)$^{\ast}$  & \textbf{75.88} $\pm$ 0.07&43.79 $\pm$ 0.11& 48.69 $\pm$ 0.04 & 43.17 $\pm$ 0.08\\
 \hline
\end{tabular}
\end{center}
\caption{Accuracy (\%) (mean and standard deviation) of different methods on ImageNet under various attacks and without attack. $^{\ast}$Average of five runs.
}
 \label{tab:title3} 
\end{table*}

\begin{table*}[t]
\begin{center}
\begin{tabular}{|l||c|c|c|c|}
\hline
Method & Without attack & PGD (100)  & FGSM & C\&W\\
  \hline
  \hline
RobNet-large~\cite{guo2020meets} & 90.73 & 87.28 & 89.43 & 69.38\\
\hline

SDARTS-ADV~\cite{chen2020stabilizing}  $^{\ast}$ & 99.19 $\pm$ 0.01 & 97.31 $\pm$ 0.02& 98.67 $\pm$ 0.02 & 78.94 $\pm$ 0.05\\
PC-DARTS-ADV~\cite{chen2020stabilizing}  $^{\ast}$ & 99.21 $\pm$ 0.01 & 97.33 $\pm$ 0.04& 98.75 $\pm$ 0.01 & 78.93 $\pm$ 0.03\\
\hline
DSRNA-CB (ours)$^{\ast}$ & 99.21 $\pm$ 0.03 & \textbf{97.34 $\pm$ 0.06}& \textbf{98.85 $\pm$ 0.03} & 94.02 $\pm$ 0.08\\
DSRNA-Jacobian (ours)$^{\ast}$ & \textbf{99.36 $\pm$ 0.01}&96.82 $\pm$ 0.02& 98.79 $\pm$ 0.01& \textbf{95.37 $\pm$ 0.02}\\
 \hline
\end{tabular}
\end{center}
\caption{Accuracy (\%) (mean and standard deviation) of different methods on MNIST under various attacks and without attack. $^{\ast}$Average of five runs.
  }
 \label{tab:title6} 
\end{table*}

\section{Experiments}
\label{sec:4}
\subsection{Dataset}

We used three datasets in the experiments: CIFAR-10~\cite{Krizhevsky09learningmultiple}, ImageNet~\cite{imagenet_cvpr09}, and MNIST~\cite{lecun-mnisthandwrittendigit-2010}. 
CIFAR-10 contains 60K images with a size  of $32 \times 32$. We use 60K images for training and the rest for validation. 
ImageNet has 1.3M training images and 50K validation images. 
MNIST has a training set of 60,000 examples and a test set of 10,000 examples, which are $28\times28$  gray-scale images of handwritten single digits between 0 and 9.

\subsection{Experimental Settings}

\subsubsection{Baselines}
We compare with the following baselines: 1) RobNet \cite{guo2020meets} which searches robust architectures based on adversarial training in one-shot NAS; 2) SDARTS-ADV and PC-DARTS-ADV \cite{chen2020stabilizing}, which performs adversarial training on architecture variables in DARTS-based NAS. We select three popular white-box untargeted adversarial attack methods to evaluate the robustness of our methods: fast gradient sign method (FGSM)~\cite{goodfellow2014explaining} , projected gradient descent (PGD)~\cite{madry2017towards}, and Carlini \& Wagner (C\&W) \cite{carlini2017towards}.

\subsubsection{Hyperparameter Settings}

The search space of our methods is the same as that of PC-DARTS,  which is composed of $3\times3$ and $5\times5$ separable convolutions, $3\times3$ and $5\times5$ dilated separable convolutions, $3\times3$ max pooling, $3\times3$ average pooling, identity, and zero. The convolutional cell consists of 6 nodes, which  has 2 input nodes, 3 intermediate nodes, and 1 output node. For CIFAR-10 and MNIST, our methods search the architectures from scratch. 
In the searching phase, a small network of 8 cells was trained for 50 epochs with an initial number of channels of 16.

In DSRNA-CB, we used SGD for optimizing the network weights $w$ with a learning rate of 0.1, a batch size of 256,  a momentum of 0.9, and a weight decay of $3e-4$. We used  the Adam optimizer \cite{kingma2014adam} for optimizing architecture variables $\alpha$, with a fixed learning rate of  $6e-4$, $\beta_1 = 0.5$, $\beta_2 = 0.999$, and a weight decay of $3e-4$. In DSRNA-Jacobian,  the network weights $w$ were optimized via SGD with a learning rate of 0.025, a batch size of 128, a momentum of 0.9,  and a  weight decay of $3e-4$. The architecture variables $\alpha$ were  optimized using  Adam~\cite{kingma2014adam} with a learning rate of $3e-4$, $\beta_1 = 0.5$, $\beta_2 = 0.999$, and a weight decay of $1e-3$.

Given the searched cell, we stack 20 copies of them into a larger network and train this network from scratch on CIFAR-10 or MNIST. 

The network was trained for 600 epochs from scratch with a batch size of 128, an initial learning rate of 0.025, norm gradient clipping of 5, drop-path with a rate of 0.3, and an initial number of channels of 36. 
For ImageNet, the architecture is transferred from CIFAR-10: given the optimal cell searched on CIFAR-10, we stack 14 copies of them into a larger network with 48 initial channels and train this network on ImageNet. The training was performed for 250 epochs using an SGD optimizer with an annealing learning rate of 0.5, a momentum of 0.9, and a weight decay of $3e-5$. The tradeoff parameter $\gamma$ in both  DSRNA-CB and DSRNA-Jacobian  was set to 0.01. 
In DSRNA-CB, we initialized $\epsilon$ as 0.03, and then linearly increased or decreased it based on the global difference between the certified upper bound and lower bound. The hyperparameters of baseline methods are deferred to the supplements.  A single NVIDIA GTX 1080Ti GPU was used to perform the search.

\begin{table*}[h]
\small
\begin{center}
\begin{tabular}{|l||c|c|c|c|c|}
 \hline
&RobNet-large~\cite{guo2020meets}&SDARTS-ADV~\cite{chen2020stabilizing}&PC-DARTS-ADV~\cite{chen2020stabilizing} & DSRNA-CB (ours)& DSRNA-Jacobian (ours)\\
 \hline
 \hline
MNIST & 0.0325 &  0.0471 &  0.0474 & \textbf{0.0526} &  0.0514  \\
 \hline
CIFAR-10 & 0.0024 & 0.0039 &   0.0040 &  \textbf{0.0049}  & 0.0048 \\
 \hline
 \end{tabular}
 \end{center}
 \caption{ Comparison of averaged $l_{\infty}$-norm  certified lower bounds of architectures searched by various methods. Larger is better.} \label{tab:title5}
\end{table*}

\begin{table*}[h]
\small
\begin{center}
\begin{tabular}{|l||c|c|c|c|c|}
 \hline
&RobNet-large~\cite{guo2020meets}&SDARTS-ADV~\cite{chen2020stabilizing}&PC-DARTS-ADV~\cite{chen2020stabilizing} & DSRNA-CB (ours)& DSRNA-Jacobian (ours)\\
 \hline
 \hline
MNIST &  0.1340 &  0.1767 & 0.1765  & \textbf{0.4288} &  0.4285\\
 \hline
CIFAR-10 & 0.0167 &  0.0337 &   0.0336 & \textbf{0.0412} &  0.0409\\
 \hline
 \end{tabular}
 \end{center}
 \caption{ Comparison of averaged $l_{2}$-norm certified lower bounds  of architectures searched by various methods. Larger is better.} \label{tab:title7}
 \vspace{-0.3cm}
\end{table*}

\subsection{Results}
In this section, we perform game-based and verification-based  evaluations of the adversarial robustness of our proposed methods and compare with  state-of-the-art baselines.
\subsubsection{Game-based Evaluation} 
Game-based evaluation estimates the success rate of defending against adversarial attacks with various forms of norm-bounds, such as  $l_{2}$, $l_{\infty}$, etc. FGSM \cite{goodfellow2014explaining,wong2020fast} and PGD \cite{madry2017towards}  are two effective $l_{\infty}$ attack methods. 
C\&W \cite{carlini2017towards} is an effective $l_{2}$ attack method.  On CIFAR-10, we evaluate our proposed methods against 1) FGSM attack with $\epsilon = 0.01$ ($2/255$), 2) PGD attack with $\epsilon = 0.03$ ($8/255$), attack iterations of 10, 20, and 100, and a step size of $2/255$, 3) C\&W with 60 attack iterations.

Table~\ref{tab:title2} shows the accuracy of different methods under various norm-bound attacks on CIFAR-10. PGD ($n$) denotes the PGD attack with $n$ iterations. From this table, we make the following observations. \textbf{First}, the accuracy of our proposed methods, including DSRNA-CB and DSRNA-Jacobian is much higher than that of other robust NAS baselines including RobNet-large, RobNet-free, SDARTS-ADV, and PC-DARTS-ADV, under PGD, FGSM, and C\&W attacks. This demonstrates that our methods are more robust against various attacks than these baselines. One major reason is that our methods search for robust architectures by explicitly and directly maximizing differentiable robustness metrics and therefore are guaranteed to obtain robust architectures. In contrast, the baseline methods try to improve the robustness of searched architectures implicitly and indirectly: performing adversarial training and injecting random noise. The implicitness and indirectness of these methods do not guarantee robustness. 
 \textbf{Second}, among the baselines, there is no consistent winner: SDARTS-ADV and PC-DARTS-ADV perform better than the other baselines under PGD and FGSM attacks; RobNet-large and RobNet-free perform better than the other baselines on C\&W attacks. None of these baselines consistently outperforms other baselines across all three types of attacks. In contrast, our proposed methods are consistently more robust than these baselines under all types of attacks.
\textbf{Third}, between our two proposed methods, DSRNA-CB is slightly more robust than DSRNA-Jacobian. This is probably because the first-order Taylor approximation in DSRNA-Jacobian incurs larger inexactness. However, DSRNA-Jacobian is much faster to train and more memory efficient than DSRNA-CB, as we will show later.

While our methods are robust against different attacks, we also would like them to be accurate when there are no attacks. To verify this, we compare the accuracy of our methods with state-of-the-art baselines under the attack-free setting. Table~\ref{tab:title1} shows the accuracy achieved by different methods on CIFAR-10 when there are no attacks. From this table, we make the following observations. \textbf{First}, the accuracy achieved by our methods are very close to the best accuracy achieved by ASAP. This demonstrates that not only being robust, our methods are also highly accurate when there are no attacks. \textbf{Second}, the accuracy of RobNet is much lower than that of ours. This shows that while our methods are not only more robust than RobNet when there are attacks, but also are much more accurate than RobNet when there are no attacks. \textbf{Third}, the search cost of our methods is similar to that of other gradient-based baselines. This demonstrates that our methods gain robustness without significantly increasing search cost. \textbf{Fourth}, while  SDARTS-ADV and PC-DARTS-ADV can  achieve high performance when there are no attacks, they are not as robust as our methods in the presence of attacks, as shown in Table~\ref{tab:title2}.

 To investigate our methods' transferability, we use the best cell structure searched on CIFAR-10  to compose a larger network and train it on ImageNet. Table \ref{tab:title3} shows the accuracy of different methods achieved on ImageNet under various norm-bound attacks and without attack. From this table, we make the following observations. \textbf{First}, under all the attacks, our methods achieve much higher accuracy than RobNet. Under C\&W attacks, our methods achieve substantially higher accuracy than SDARTS-ADV and PC-DARTS-ADV. Under PGD and FGSM attacks, our methods are on par with SDARTS-ADV and PC-DARTS-ADV: our methods are slightly better than SDARTS-ADV and PC-DARTS-ADV under FGSM attacks;  SDARTS-ADV and PC-DARTS-ADV are slightly better than our methods under PGD attacks. These results further demonstrate that our methods are more robust against various types of attacks than the baselines. \textbf{Second}, when there are no attacks, the accuracy of our methods is much higher than that of RobNet. In addition to being more robust, our methods are also more accurate than RobNet under the attack-free setting. \textbf{Third}, DSRNA-CB is slightly more robust than DSRNA-Jacobian.

Table \ref{tab:title6} shows the results on MNIST. Similarly, our methods are substantially more robust than RobNet-large under all types of attacks, and are substantially more robust than SDARTS-ADV and PC-DARTS-ADV under C\&W attacks. Our methods are on par with SDARTS-ADV and PC-DARTS-ADV under PGD and FGSM attacks. When there is no attack, our methods achieve much higher accuracy than  RobNet-large and are on par with SDARTS-ADV and PC-DARTS-ADV.

\vspace{-0.3cm}
\paragraph{Runtime} With a single GTX 1080Ti GPU, the runtime on CIFAR-10 for the search phase of DSRNA-CB is 4 GPU days, while that of DSRNA-Jacobian is 0.4 GPU days. 
On MNIST, DSRNA-CB takes 1 GPU day for architecture search while DSRNA-Jacobian takes 0.2 GPU days. DSRNA-Jacobian is more efficient than DSRNA-CB, but is less robust than DSRNA-CB as shown previously.

\vspace{-0.3cm}
\subsubsection{Verification-based Evaluation}
In this section, we use the certification method developed in Section \ref{sec:cb} to find the certified lower bounds of the architectures searched by different  methods. Larger lower bound indicates  more robustness.  Table~\ref{tab:title5} and Table~\ref{tab:title7} compare the averaged certified lower bounds of architectures searched by different methods on MNIST and CIFAR-10 under $l_{2}$ and $l_{\infty}$ norms. As can be seen, the lower bounds achieved by our methods under various norms are larger than those achieved by baselines. This further demonstrates that our methods are more robust than these baseline methods.

\section{Conclusion}
\label{sec:5}
To address the problem that existing neural architecture search (NAS) methods are vulnerable to adversarial attacks, we propose methods for differentiable search of robust architectures. We define two differentiable measures of architectures' robustness, based on certified robustness lower bound and Jacobian norm bound. Then we search for robust architectures by performing optimization in the architecture space with an objective of maximizing the robustness metrics. On various datasets, we demonstrate that our methods 1) are more robust to various norm-bound attacks than several robust NAS baselines; 2) are more accurate than baselines when there are no attacks; 3) have significantly higher certified lower bounds than baselines.

\typeout{}

{\small
\bibliographystyle{ieee_fullname}
\bibliography{Paper}
}

\end{document}